%% file: main-DeHiDe.tex
\documentclass[runningheads]{llncs}

\usepackage{booktabs} 
\usepackage{xspace}
\usepackage{mathtools}
\usepackage{amsmath,amssymb,latexsym}
\usepackage{enumitem}
\usepackage{algorithm}
\usepackage{algorithmicx}
\usepackage[noend]{algpseudocode}
\usepackage{graphicx} 
\usepackage{color}
\usepackage{caption}
\usepackage{tabularx}
\usepackage{multicol}
\usepackage{float}
\usepackage{xcolor}
\usepackage[justification=centering]{caption}
\usepackage{todonotes}
\usepackage{lipsum}
\usepackage{tabularx}
\newcolumntype{L}{>{\raggedright\arraybackslash}X}
\usepackage{subfig}
\usepackage{soul}

\input{macros}

\begin{document}

\title{DeHiDe: \underline{De}ep Learning based \underline{H}ybr\underline{i}d Model to \underline{De}tect Fake News using Blockchain\thanks{An original proposal of this work is accepted at the Doctoral Symposium ICDCN 2021~\cite{Agarwal+Poster:ICDCN:2021}.
}}	


\author{Prashansa Agrawal \and Parwat Singh Anjana \and Sathya Peri\thanks{Author sequence follows lexical order of last names.}}
\institute{Department of CSE, Indian Institute of Technology, Hyderabad, India\\ \email{\texttt{cs18mtech11037@iith.ac.in, cs17resch11004@iith.ac.in, sathya\_p@cse.iith.ac.in}}}
\authorrunning{Agrawal et al.}
\titlerunning{\underline{De}ep Learning based \underline{H}ybr\underline{i}d Model to \underline{De}tect Fake News using Blockchain}

\maketitle
\begin{abstract}
The surge in the spread of misleading information, lies, propaganda, and false facts, frequently known as fake news, raised questions concerning social media's influence in today's fast-moving democratic society. The widespread and rapid dissemination of fake news cost us in many ways. For example, individual or societal costs by hampering elections integrity, significant economic losses by impacting stock markets, or increases the risk to national security. It is challenging to overcome the spreading of fake news problems in traditional centralized systems. However, Blockchain-- a distributed decentralized technology that ensures data provenance, authenticity, and traceability by providing a transparent, immutable, and verifiable transaction records can help in detecting and contending fake news. This paper proposes a novel hybrid model \emph{DeHiDe: \underline{De}ep Learning-based \underline{H}ybr\underline{i}d Model to \underline{De}tect Fake News using Blockchain}. The DeHiDe is a blockchain-based framework for legitimate news sharing by filtering out the fake news. It combines the benefit of blockchain with an intelligent deep learning model to reinforce robustness and accuracy in combating fake news's hurdle. It also compares the proposed method to existing state-of-the-art methods. The DeHiDe is expected to outperform state-of-the-art approaches in terms of services, features, and performance.

\end{abstract}

\input{intro}

\input{rw}
\input{approach}

\input{architecture}
\input{results}

\input{conclusion}

\bibliographystyle{splncs04}
\bibliography{citations}

\end{document}

%% file: macros.tex
\newcommand{\punt}[1]{}
\newcommand{\cmnt}[1]{}

\newcommand{\secref}[1]{Section~\ref{sec:#1}}
\newcommand{\figref}[1]{Fig.~\ref{fig:#1}}
\newcommand{\tabref}[1]{Table~\ref{tab:#1}}

\newcommand{\eqnref}[1]{Eqn(\ref{eq:#1})}

\newcounter{linenumber}

\newcommand{\remove}[1]{}

\newcommand{\ignore}[1]{}

\newcommand{\bc} {blockchain\xspace}

\algrenewcommand{\algorithmiccomment}[1]{$//$ #1}

\algrenewcommand\alglinenumber[1]{\tiny #1:}
\algnewcommand\algorithmicswitch{\textbf{switch}}
\algnewcommand\algorithmiccase{\textbf{case}}
\algnewcommand\algorithmicassert{\texttt{assert}}
\algnewcommand\Assert[1]{\State \algorithmicassert(#1)}%
\algdef{SE}[SWITCH]{Switch}{EndSwitch}[1]{\algorithmicswitch\ #1\ \algorithmicdo}{\algorithmicend\ \algorithmicswitch}%
\algdef{SE}[CASE]{Case}{EndCase}[1]{\algorithmiccase\ #1}{\algorithmicend\ \algorithmiccase}%
\algtext*{EndSwitch}%
\algtext*{EndCase}%

\newcommand{\bcheck}{{\color{black}\checkmark}}

%% file: intro.tex
\section{Introduction}
\label{sec:intro}
In recent years, the Internet made it too easy to obtain the latest news in milliseconds; however, this advancement also leads to digital deceptions and misleading fake news. Fake news is being propagated on various platforms to present misinformation or hoaxes. With deceptive words, online social network users can easily get infected with fake news, which has already had a considerable impact on offline society. For example, the prevalence of fake news on various platforms like news channels, streaming networks, and social media elevates the wrong opinions amongst individuals and society as well as endangers national security and democracy. It also hinders elections integrity and influences stock markets, leading to significant economic losses~\cite{FragaDigitalDeception2020}. During the 2016 US presidential election, different kinds of fake news disseminated widely through online social networks that directly impacted election outcomes. According to the post-election statistical survey, online social networks account for more than 41.8\% of false news traffic during the election, much higher than the data traffic shared on other platforms.

Further, due to barrierless platforms being present for publishing the news, traditional centralized systems are inefficient in controlling and stopping fake news. Moreover, due to a central controlling entity like a news agency, it is impossible to stop the spreading of misleading information. In such situations, building robust decentralized systems becomes very crucial.

Blockchain provides a spectrum of desirable syntactic and semantic properties such as provenance, immutability, verifiability, and data integrity. It offers transparent, immutable, and verifiable records in the form of transaction blocks linked with a cryptographic hash stored in a decentralized distributed public network~\cite{FragaDigitalDeception2020}. A \bc network consists of multiple peers (or nodes) where peers do not necessarily trust each other. Each node maintains a copy of the distributed ledger. \emph{Clients}, users of the \bc, send requests or \emph{transactions} to the nodes of the \bc called as \emph{miners}. The miners collect multiple transactions from clients to form a \emph{block}. Multiple miners propose new blocks to be added to the distributed ledger. Peers follow a global consensus protocol to agree on which blocks are chosen to be added and in what order. While adding a block to the \bc, the miner incorporates the previous block's hash into the current block. This makes it difficult to tamper with the distributed ledger. The resulting structure is in the form of a linked list or a chain of blocks, hence the name \bc~\cite{anjana:ObjSC:Netys:2020,Anjana+:CESC:PDP:2019}. 

These characteristics of blockchain technology make it an ideal choice for deploying a security system to exchange credible news by detecting and preventing fake news dissemination. However, using the blockchain would not be enough to prevent fake news. We need to follow the traditional classical method of detecting and recognizing false news. Additionally, we also need to give importance to news publishers and consumers. These can be achieved by special intelligent algorithms, such as deep learning models, that can also be consolidated to extend the system's robustness by classifying the news precisely as false or real.

Text classification has an extensive scientific history and many useful applications within the natural language processing community to prove its significance. While there are various resources and tools to recognize fake news sources, for example, if a website or an individual publishes fake news, this issue can be addressed as an instance of text classification, using news content based on title, body, and source as features. Doing so helps in developing deep learning models to detect fake news. When provided as an input, a deep learning model can use text classification to predict an article's nature based on the publisher's credibility and history. Further, to create a vector space of words and develop a lingual relationship, text embedding and pre-processing can be performed. In the existing literature, benchmark results to predict fake news have been obtained by combining the convolution neural network and recurrent neural network—the usefulness of word embedding complements the model altogether \cite{agarwal2020fake}.

Different model parameters tuning and periodically training the model can ensure higher prediction efficiency for the best outcomes. Nevertheless, existing solutions using deep learning models can detect fake news but fails to deter fake news dissemination by malicious reporters. In contrast, this paper blends the advantages of blockchain with deep learning models to detect and prevent fake news dissemination in mass media.

\noindent
\textbf{Contributions: }The significant contributions of the proposed framework are as follows:
\begin{itemize}
    \item We proposed a novel hybrid model called \emph{DeHiDe: \underline{De}ep Learning-based \underline{H}ybr\underline{i}d Model to \underline{De}tect Fake News using Blockchain}. DeHiDe is a blockchain-bases framework for legitimate news sharing by filtering out the fake news. 
    \item The proposed approach is a hybrid approach that combines the benefit of blockchain with an intelligent deep learning model. 
    \item In DeHiDe, blockchain provides provenance, traceability, and security, while the deep learning model reinforces accuracy in identifying fake news based on historical observations and facts.
    \item We compare the proposed approach with state-of-the-art methods in terms of services, features, and performance. The DeHiDe is expected to outperform existing approaches.
\end{itemize}

The rest of the paper is organized as follows: the \secref{rw} presents related work closely in line with the proposed approach. A detailed explanation of the proposed methodology is given in \secref{sa}, while \secref{psa} provides detailed dissipation of various components of the DeHiDe framework. 
The \secref{results} talks about the evaluation of models in terms of features. Finally, in \secref{conc}, we will conclude with some future research directions.

%% file: rw.tex
\section{Related Work}
\label{sec:rw}
In the literature, researchers have done a significant amount of work to detect fake news using deep learning models; however, detection and prevention of fake news using blockchain technology is not adequately studied; moreover, no research combines the benefit of both to counter the fake news problem. This section presents a summary of the most relevant work in line with the proposed approach.

Owing to the successful growth of online social networks, fake news for various commercial and political benefits exists in large numbers. Several research studies have been conducted to identify the source of fake news or evaluate the news text using Natural Language Processing (NLP) models to classify news in two categories as fake or real. Existing approaches identified fake news using various Deep Learning models, including Logistic Regression (LR), Feed-forward Network, Gated Recurrent Units (GRUs), long-term memory (LSTMs), max-pooling Convolutional Neural Network (CNN), and Max-pooling and Attention CNN.

Zhang et al.~\cite{ZhangFakeDetector2020} proposed a novel gated graph neural network which aims to identify fake creators and subjects along with fake news articles. Agarwal et al.~\cite{agarwal2020fake} experimented with GloVe (Word Representation Global Vectors) word embedding to map words in the feature space and establish lingual relationships. The authors proposed merging CNN and Recurrent Neural Networks (RNN) architecture claimed to outperform gated recurrent units and RNN. Bajaj~\cite{Bajaj2017TP} proposed a novel design that integrates an attention-like mechanism into the convolutional network that can predict whether a piece of news is fake based on news content and compares the results of several different natural language processing models. The results are compared on the basis of \emph{accuracy, recall}, and \emph{F1 score}, with results showing that the GRU model outperforming in terms of \emph{recall} and \emph{F1 score} while attention-based CNN showed the best \emph{accuracy}.

Standalone systems with artificial intelligence support are prone to exploitation in terms of integrity and correctness. Besides, due to barrierless centralized platforms (such as news agencies) being present for publishing the news, it is impossible to stop spreading fake information. Hence, the deep learning architectures presented in the existing studies need a strong immutable platform to maintain system integrity and make it tamper-proof. In such situations, building robust decentralized systems becomes very crucial. The blockchain technology can be a suitable candidate for implementing a secure and reliable system.
{
}

The potential of blockchain in the battle against fake news is not studied adequately in the literature. Existing approaches either concentrate on tracing fake news sources or preventing a potentially malicious individual from publishing. However, no solution classifies news in real-time and prevents fake news from spreading in the public domain. 

Arquam et al.~\cite{arquam2018blockchain} proposed a technique to identify fake news where users are assigned credits and two types of trust--global and local, to reflect user's overall and mutual trust. Balouchestan et al.~\cite{balouchestani2019sanub} proposed a fake news detection model using blockchain in which three different profiles of users as reporters, analyzers, and validators are presented. A user can publish news as a reporter, rate the news, work as an analyzer, and validate it as a validator. Chen et al.~\cite{chen2018towards} suggested a blockchain network using an improved Susceptible, Infectious, Recovered model to evaluate and control rumors and fake news. Islam et al.~\cite{islam2020newstradcoin} proposed a platform for publishers to collect news securely from reporters without a third party interference for data integrity. Paul et al.~\cite{paul2019fake} proposed a fake news detection model using blockchains. Qayyum et al.~\cite{qayyum2019using} generate a chain of news that does not rely on the news; instead, the sources and publishers. However, news agencies can publish fake news. Shang et al.~\cite{shang2018tracing} proposed a system containing the original news chain to prevent the news tempering. However, publishers themselves may publish fake news that can not be prevented.

In essence, the classical NLP based model can classify the news and source of the news as fake or real but can not stop the spread of fake news in real-time. While existing blockchain-based approaches can trace fake news's origin or not allow a malicious reporter to publish news. There is no real-time solution to stop and filter the fake news from being published. We proposed a hybrid technique that combines the best of evolutionary blockchain technology and classical deep learning models to overcomes the limitations of individual approaches. In DeHiDe, the dynamically changing credibility is adopted from~\cite{balouchestani2019sanub}, while the idea of selecting people belonging to the same area as analyzers is selected from~\cite{paul2019fake}. The deep learning models to classify fake news is selected from the work in~\cite{agarwal2020fake}. So in the proposed approach, blockchain provides provenance, traceability, and security, while the deep learning model reinforces accuracy in classifying fake news based on historical observations and facts.

%% file: approach.tex
\section{Solution Approach}
\label{sec:sa}

This section presents a high-level sketch of the proposed solution.

Practical applications to detect fake news could include testing the news source, scrapping web data, and finding which outlets are more likely to spread fake news. However, in the absence of such source tests, it is determined that looking at common linguistic features in the source articles, including sentiment, sophistication, and layout, is the most reliable way to detect false news. For example, fake news agencies were more likely to use hyperbolic, subjective, and emotional language. In the proposed approach, similar to~\cite{agarwal2020fake}, using pre-trained GloVe embedding matrix, the text is pre-processed and rendered to the embedding layer. Besides, coevolutionary layers, max-pooling layers, and LSTM layers are the paradigm of the proposed methodology. By observing and analyzing linguistic features, the model is trained and used to predict the news's essence.

We propose a novel DeHiDe architecture that incorporates blockchain as a secure decentralized storage platform for sharing trustworthy news, further detecting and filtering fake news based on historical facts using a strong deep learning model. In DeHiDe, any user can publish the news as a reporter along with supporting evidence. The randomly selected analyzers analyze the news using a deep learning model. The news then presented to verifiers based on the news's locality of origin. The verifiers are actual news readers who give news ratings based on their creed. DeHiDe also dynamically changes the user's credibility to penalize reporters and analyzers if they propagate fake news.

%% file: architecture.tex
\section{DeHiDe Architecture} 
\label{sec:psa}
\begin{figure}[!t]
	\centering
	\includegraphics[width=.9\columnwidth]{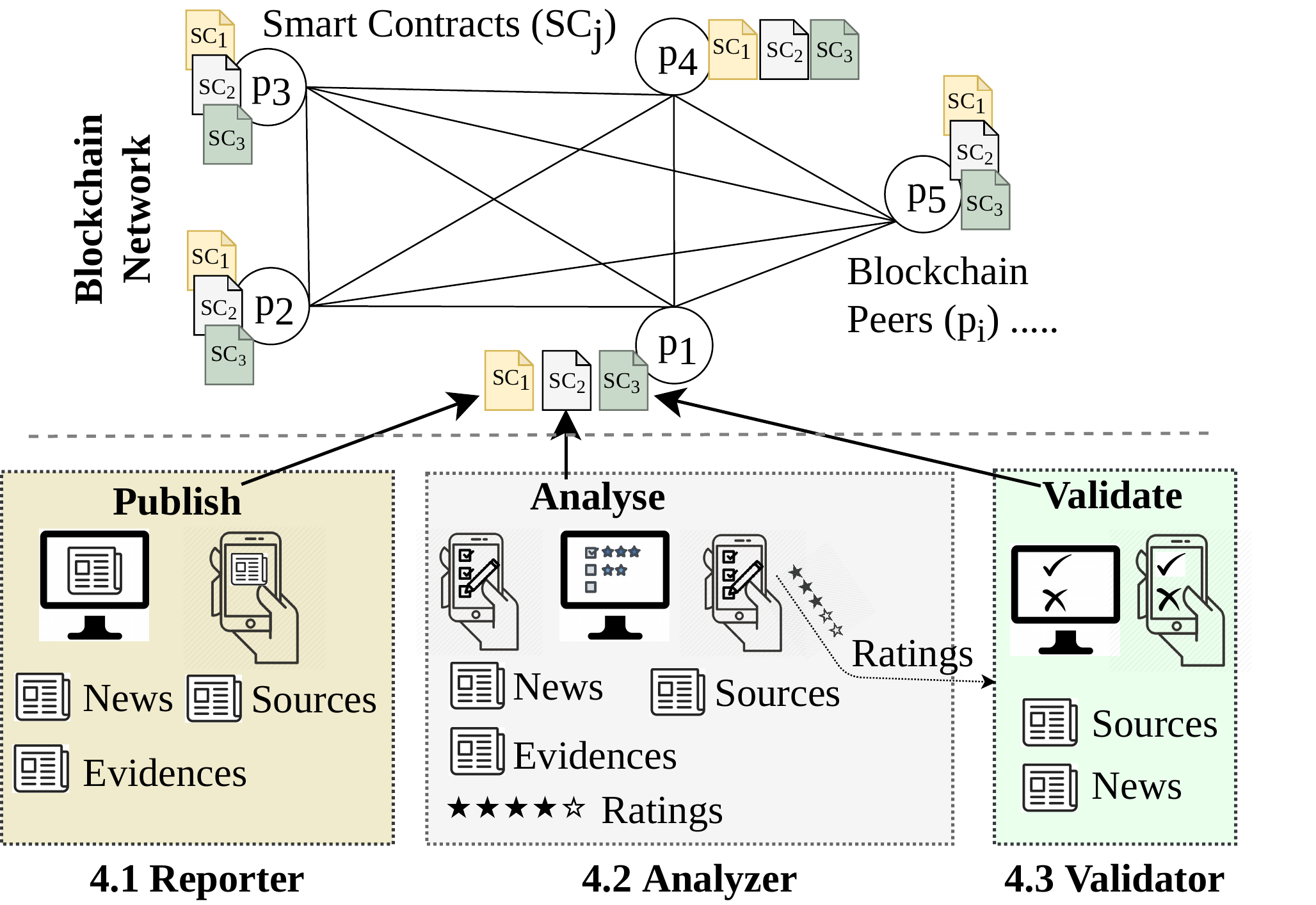}\vspace{-.1cm}
	\caption{A high-level overview of DeHiDe architecture.}
	\label{fig:DeHiDe}
\end{figure}

This section provides a high-level overview of the proposed DeHiDe architecture. As shown in the {\figref{DeHiDe}}, a user can have three different system profiles: Reporter, Analyzer, and Validator. All three entities are bound to use smart contracts using respective distributed applications. The ratings given by analyzers, validators, reporters, and other relevant information are stored as transactions in the blockchain. Since in DeHiDe, anyone can publish the news, and others validate the authenticity of the news. Therefore blockchain provides a distributed, trusted, secure storage to store every action taken by the users, which is used as proof—further used in the deep learning model to detect fake news.

\subsection{Reporter} 
When publishing news, a user is known as a reporter who can anonymously publish news and supporting evidence using the reporter's application. A reporter's identity is not disclosed until the news gets published for a fair analysis of the news by the analyzers and validators. A reporter provides various tags concerning the news topic, source, and news domain to help the system choose analyzers and reach out to validators. A reporter's reputation is defined using credibility ($\delta_r$) that varies from $0$ to $1$ and set to $0.5$ at the very outset of the process. The reporter's credibility increases when the news gets published. Unlike literature, we propose that reporters assign a specific news rating that defines their belief. A reporter gives its rating ($\mathcal{R}$) using \eqnref{reporter-rating}.


\begin{equation}
	\label{eq:reporter-rating}
	\mathcal{R} = \lambda_{r} * \delta_{r}
\end{equation}
Where $\lambda_{r}$ is rating given by reporter ($0 \leqslant \lambda_{r} \leqslant 5$), $\delta_r$ is credibility of the reporter ($0 \leqslant \delta_r \leqslant 1$) as defined in \eqnref{reporterCredRating}.

\subsection{Analyzer}
Analyzers analyze the news and give the rating through analyzer application. Analyzers verify the news using supporting news evidence, sources, and data stored in the blockchain. Initially, all the analyzers' credibility is set to $0.5$, which may increase or decrease based on their news assessment. The \eqnref{analyzer-rating} is the formulation of the analyzer's overall rating.

\begin{figure}[!t]
	\centering
	\includegraphics[width=.75\columnwidth]{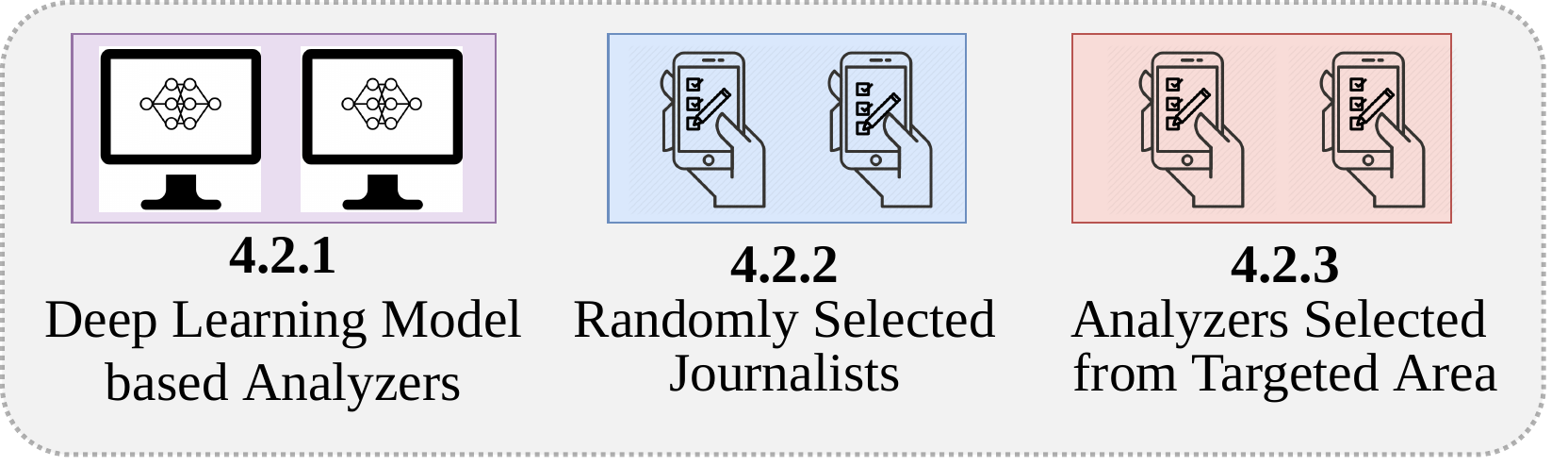}\vspace{-.3cm}
	\caption{Different types of analyzers in DeHiDe.}
	\label{fig:Analyzer}
\end{figure}
\begin{equation}
	\label{eq:analyzer-rating}
	\mathcal{A} = \frac{1}{n}\sum_{i=1}^{n}\delta_{ai}*\lambda_{ai}
\end{equation}
Where $\delta_{ai}$ is credibility of the analyzer $i$ ($0 \leqslant \delta_{ai} \leqslant 1$), $\lambda_{ai}$ is rating by the analyzer $i$ ($0 \leqslant \lambda_{ai} \leqslant 5$), $n$ is the total number of randomly selected analyzers. As shown in \figref{Analyzer}, there are three types of analyzer as follows:

\vspace{.4cm}
\noindent
\textit{4.2.1 Deep Learning Models based Analyzer}
\vspace{.1cm}

\begin{figure}[H]
\centering
	\includegraphics[width=.8\columnwidth]{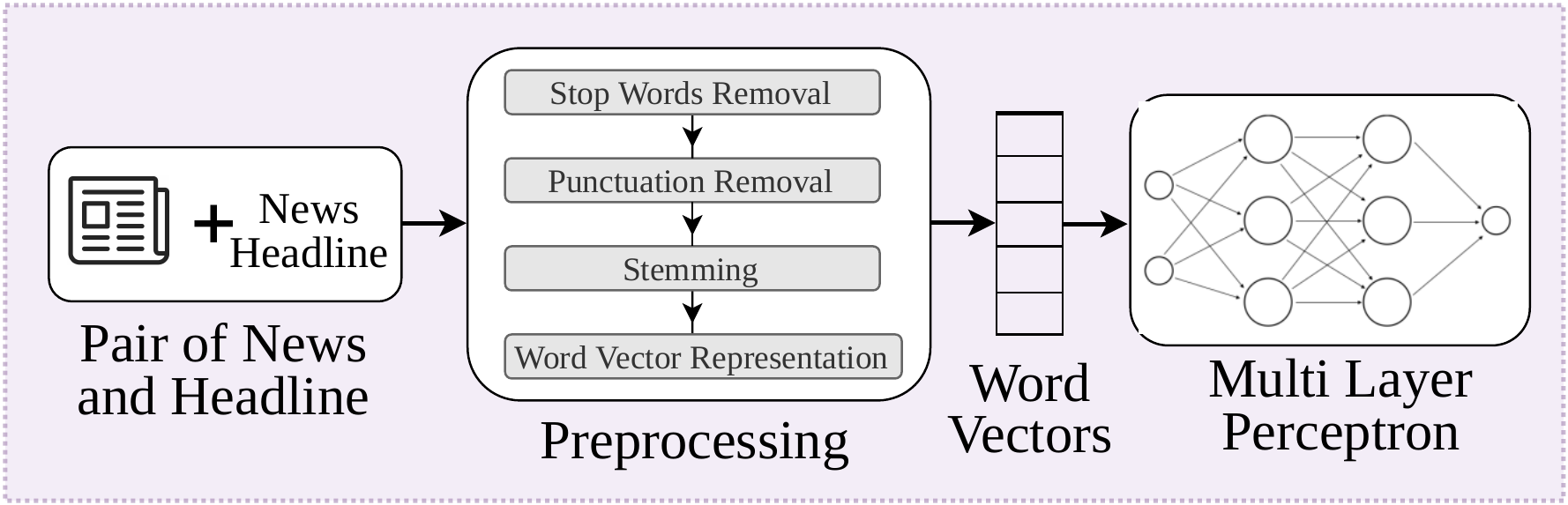}
	\vspace{-.3cm}
	\caption{A high-level overview of deep learning model for fake news detection.}
	\label{fig:DLModel}
\end{figure}

\begin{figure}[t]
\centering
	{\includegraphics[width=.42\columnwidth]{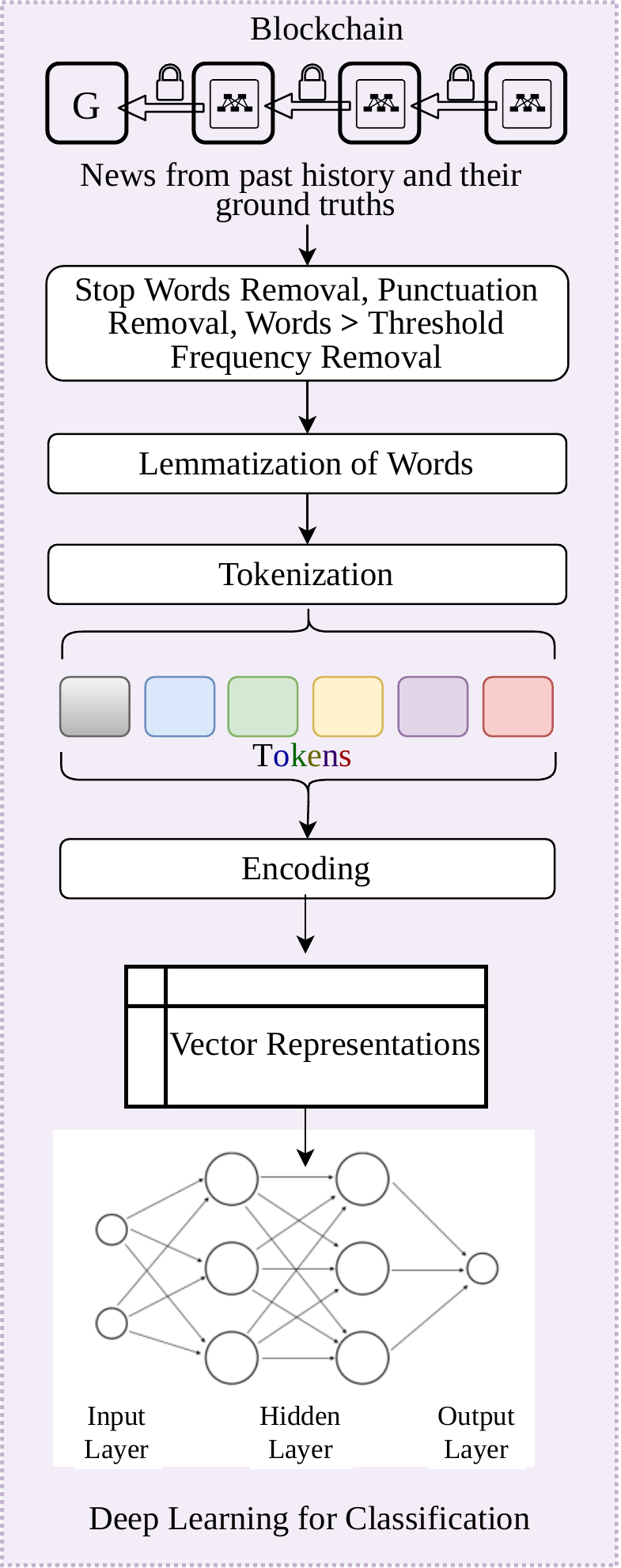}}\vspace{-.25cm}
	\caption{Information flow in deep learning model.}
	\label{fig:DLFlow}
\end{figure}

\begin{figure}[t]
\centering
	{\includegraphics[width=.55\columnwidth]{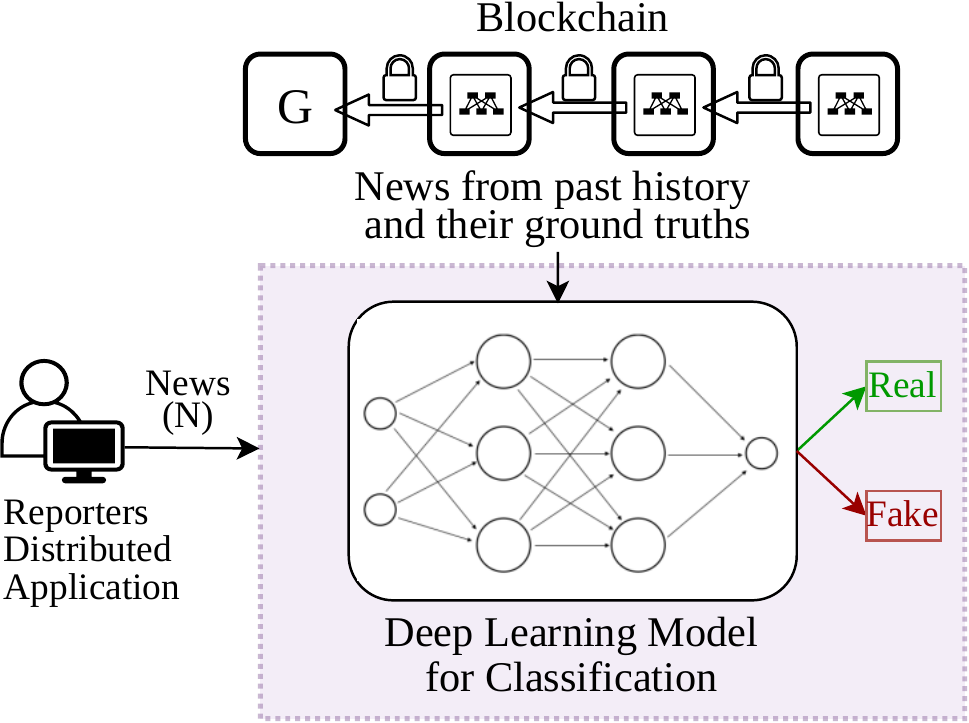}}\vspace{-.3cm}
	\caption{Deep learning model-based analyzer in DeHiDe.}
	\label{fig:DLAnalyzer}
\end{figure}

\noindent
The deep learning (DL) model based analyzer is a principal component of the proposed model and helps achieve the better classification of news published by the anonymous reporters. It tackles the problem of detecting fake news from a natural language processing perspective. The DL-based analyzers perform the classification of articles into two types fake or real, based on the earlier published news data in the blockchain. The DL model does not consider reporters' identities for news classification since reporters' identities are not disclosed until the news gets published. A high-level overview of deep learning model based analyzer is shown in \figref{DLModel} to \figref{DLAnalyzer}.


As shown in \figref{DLFlow}, pre-processing of the data is done in the first phase. Different techniques are used to keep the important words in the data and remove the words of lesser significance in classification. Stop words, extra symbols, and words that occur more than the threshold are removed for this purpose. Further lemmatization is carried out in order to bring all the words to their basic forms called \emph{lemma}. For example, words like ``see'' and ``saw'' would come in the same bucket of words. 

In the next step, all the words in the sentences are to be represented in the form of word embedding. To carry out this task, various methods like GloVe, word2vec, ELMo, BERT word embedding can be used. The word representations given as input to Deep Learning Models such as GRU, CNN, LSTM, or simple Machine Learning classifiers such as Support Vector Machine (SVM), Gradient Boosting (GB), etc. Stance detection is another version of fake news detection, which essentially discerns whether the actual body of the news talks about what has been presented in the news headline—the high-level overview of stance detection is presented in the {\figref{DLFlow}}. The deep neural network is given word vectors as input for classification to detect similarity between news titles and bodies (text). 

As shown in \figref{DLAnalyzer}, the news is detected as fake based on the similarity between known facts about earlier detected fake news and historical data with evidence stored in the blockchain. If a DL-based analyzer detects the news as fake, it assigns a $0$ rating to the news; otherwise, assign a positive value less than and equal to $5$ if the news is classified as real.

\vspace{.4cm}
\noindent
\textit{4.2.2 Randomly Selected Journalists}
\vspace{.1cm}

\noindent
Users with verified journalist profiles are randomly chosen to rate the news. Random selection makes it less prone to paid reviews since any journalist may be chosen to rate the news reported by an anonymous reporter. The journalist's interpretation of the news stories can be more reliable because they have more experience of the past news. A journalist can be a reporter, too, and they know the importance of the news being reported. In the proposed approach, a reporter can also be an analyzer for other reporters. When a reporter reports news, various analyzers, including randomly chosen journalists, evaluate the news. So our news analysis not relies only on the classic deep learning model and gives importance to the journalists and the analyzer present in the places where the news originated. Besides that, reporters' identity is not disclosed before the news gets published; thus, there is no chance that the unfair review will affect the news rating.

\vspace{.4cm}
\noindent
\textit{4.2.3 Analyzers from Targeted Area}\vspace{.1cm}

\noindent
Users belonging to the same area where the incident occurred, or those belonging to the same domain, are randomly chosen based on user tags. They are more likely to grasp the news's truth since they belong to the news's origin or are experts in the domain. In the proposed approach, analyzers are selected randomly from targeted areas and domains based on the location and domain information given by the analyzers while using analyzers distributed applications. Such analyzers analyze the news and provide the rating based on their experience and close association with the source from where the news is originated. Since the reporter provides the news source along with evidence, an analyzer can quickly provide their rating. For example, suppose a reporter reported a piece of news regarding communal violence in $x$ locality. The analyzer from the locality $x$ might better understand the actual situation and reasonably analyze the news. If analyzers provide wrong analysis in support of the fake news. It will negatively impact their credibility when the news is classified as fake.

\subsection{Validator} 
Validators are actual news viewers who read the news as per their interests based on certain news tags. According to the ratings given by analyzers, news articles are presented to the validators. A news $\mathcal{N}$ is not released in the public domain and is assumed to be fraudulent if the rating falls below a certain threshold. Validators validate the news based on how they perceive it or suspect it. The validator's confidence factor is often illustrated and influences news exposure on mainstream media. The proportion of the validators creed ($\mathcal{V}$) is given in \eqnref{validator-belief}.

\begin{equation}
	\label{eq:validator-belief}
	\mathcal{V} = \frac{\eta_b}{\eta} * 5
\end{equation}
Where ${\eta_b}$ is the number of validators who believed the news, $\eta$ is the number of validators who validated the news $\mathcal{N}$. 

The total score $\mathcal{S}_{total}$ for news $\mathcal{N}$ is computed using \eqnref{overallNewsRating}. The maximum score can be $1$, while the minimum can be $0$. If the score is below threshold (e.g., $.5$), then the probability of news being fake is high since the majority (analyzer and validator) reported the news as fake.

\begin{equation}
	\label{eq:overallNewsRating}
	\mathcal{S}_{total} = \frac{(\mathcal{R} * \omega_1) + (\mathcal{A} * \omega_2) + (\mathcal{V} * \omega_3)}{5}
\end{equation}
Where $\mathcal{R}$ is reporters rating, $\mathcal{A}$ is the rating given by analyzers, $\mathcal{V}$ is validators belief in the news $\mathcal{N}$, $\omega_1$, $\omega_2$, and $\omega_3$ are constants such that $\sum_{i =1}^{3}\omega_{i} = 1$ that define the weightage of the reporter, analyzers, and validators respectively, and $5$ is the maximum rating.

\subsection{Credibility}
The system participants are assigned with initial credibility score ($0.5$) that changes between $0$ to $1$. The reporter's credibility will be significantly affected by the penalization factor ($\theta$) if publishing false news. Higher the penalization factor more significant the penalty. The formulation of reporter credibility ($\delta_r$) is given in the \eqnref{reporterCredRating}.

\begin{equation}
	\label{eq:reporterCredRating}
	\delta_r = {\frac{\delta_r}{\delta_r + (\theta * \frac{\mathcal{F}_n}{\mathcal{G}_n})} }
\end{equation}
Where, the total number of fake news and authentic news published by a reporter is $\mathcal{F}_{n}$ and $\mathcal{G}_{n}$. 

An analyzer's reputation is calculated using \eqnref{AnalyzerCred}. Suppose an analyzer falsely rates news as false (or authentic), which eventually verifies as accurate (or false), then analyzer's credibility decreases based on the number of news incorrectly evaluated by the analyzer and the confidence of validators in the analyzer for news $\mathcal{N}$.

\begin{equation}
	\label{eq:AnalyzerCred}
	\delta_a = \frac{\mathcal{T}_p + \mathcal{T}_n}{\mathcal{T}_p + \mathcal{T}_n + \mathcal{F}_p + \mathcal{F}_n} * \frac{\eta_a}{\eta}
\end{equation}
Where $\mathcal{T}_p$ is true positive, $\mathcal{T}_n$ is true negative, $\mathcal{F}_p$ is false positive, and $\mathcal{F}_n$ is false negative analysis done by the analyzer $a$. $\frac{\eta_a}{\eta}$ is a ratio of how many validators agree with the analyzer to the total number of validators who validated the news.


%% file: results.tex
\section{Discussion}
\label{sec:results}
As shown in \tabref{comp}, we compared DeHiDe based on various features with existing methods in the state-of-the-art literature. From the \tabref{comp}, it can be observed that many features that support better accuracy in detecting fake news and maintaining quality provided in the DeHiDe. In the earlier works, either of blockchain or deep learning models are used to detect fake news. We propose a novel idea combine blockchain with intelligent deep learning models in single architecture for better accuracy. We also propose the criteria to choose the evaluators in the system without bias. The proposed architecture is currently partially implemented. As future work, we plan to implement a fully functioning real-time system to detect and filter fake news.

\begin{table}[H]
\caption{DeHiDe comparison with state-of-the-art blockchain-based solutions.
}
\label{tab:comp}
\resizebox{1\columnwidth}{!}{%
        \begin{tabular}{l|c|c|c|c|c|c|c|c}
        \hline
            \textbf{Features}            &
            
            \begin{tabular}[c]{@{}c@{}}{Chen}\\{et al.~\cite{chen2018towards}} \end{tabular}  &
            
            \begin{tabular}[c]{@{}c@{}}{{Islam}}\\{{et al.~\cite{islam2020newstradcoin}}} \end{tabular} &
            
            \begin{tabular}[c]{@{}c@{}}{Qayyum}\\{et al.~\cite{qayyum2019using}}\end{tabular} &
            
            \begin{tabular}[c]{@{}c@{}}{Shang}\\{et al.~\cite{shang2018tracing}} \end{tabular}  &
            
            \begin{tabular}[c]{@{}c@{}}{Arquam}\\{et al.~\cite{arquam2018blockchain}} \end{tabular}  &
            
            \begin{tabular}[c]{@{}c@{}}{Balouchestani}\\{et al.~\cite{balouchestani2019sanub}}\end{tabular} &
            
            \begin{tabular}[c]{@{}c@{}}{Paul}\\{et al.~\cite{paul2019fake}}\end{tabular} &
            
            \textbf{DeHiDe} \\ \hline\hline
            
            \begin{tabular}[c]{@{}c@{}}\textbf{News} \textbf{Broadcasting}\end{tabular} & 
            $\bcheck$ &
            $\bcheck$ &
            $\bcheck$ &
            $\bcheck$ &
            $\bcheck$ &
            $\bcheck$ &
            $\bcheck$ &
            $\bcheck$ \\ \hline
            
            \begin{tabular}[c]{@{}c@{}}\textbf{Ownership} \textbf{Proof}\end{tabular} &
            &
            $\bcheck$ &
            $\bcheck$ &
            $\bcheck$ &
            $\bcheck$ &
            $\bcheck$ &
            &
            $\bcheck$ \\ \hline
            
            \begin{tabular}[c]{@{}c@{}}\textbf{Fake News} \textbf{Detection}\end{tabular} &
            &
            &
            $\bcheck$ &
            $\bcheck$ &
            $\bcheck$ &
            $\bcheck$ &
            $\bcheck$ &
            $\bcheck$ \\ \hline
            
            \begin{tabular}[c]{@{}c@{}}\textbf{Reporters} \textbf{Validation}\end{tabular} & 
            &
            &
            &
            &
            $\bcheck$ &
            $\bcheck$ &
            &
            $\bcheck$ \\ \hline
            
            \begin{tabular}[c]{@{}c@{}}\textbf{News} \textbf{Evaluation}\end{tabular} &
            &
            &
            &
            &
            &
            $\bcheck$ &
            $\bcheck$ &
            $\bcheck$ \\ \hline
            
            \begin{tabular}[c]{@{}c@{}}\textbf{Analyst} \textbf{Existence}\end{tabular} & 
            &
            &
            &
            &
            &
            $\bcheck$ &
            $\bcheck$ &
            $\bcheck$ \\ \hline
            
            \begin{tabular}[c]{@{}c@{}}\textbf{Native User}  \textbf{Importance}\end{tabular} &
            &
            &
            &
            &
            &
            &
            $\bcheck$ &
            $\bcheck$ \\ \hline
            
            \begin{tabular}[c]{@{}c@{}}\textbf{Random Analyst} \textbf{Selection}\end{tabular} &
            &
            &
            &
            &
            &
            &
            $\bcheck$ &
            $\bcheck$ \\ \hline
            
            \begin{tabular}[c]{@{}c@{}}\textbf{DL (NLP)} \textbf{Model}\end{tabular} &
            &
            &
            &
            &
            &
            &
            &
            $\bcheck$ \\ \hline
            
            \begin{tabular}[c]{@{}c@{}}\textbf{Real-time} \textbf{System}\end{tabular} &
            &
            &
            &
            &
            &
            &
            &
            $\bcheck$ \\ \hline

            \end{tabular}%
}
\end{table}




Our prototype model's end-to-end implementation is left as future work to test effectiveness and efficiency. We will implement DeHiDe as a smart contract on the public blockchain and test it on a real-world fake news dataset. The implementation so far focuses on Analyzer's application on the Ethereum blockchain. Each Analyzer is assigned with some initial weight; however, weights change over time depending on how honestly and correctly they analyzed and verified the news. The news ratings are calculated using the formulations discussed in \secref{psa}. The final news rating is calculated based on Reporters, Analyzers, and validator's ratings. Ratings given by analyzers are stored in the blockchain and cannot be changed—the higher the news rating, the better its visibility on the DeHiDe platform. The news falling below the threshold are not allowed to show up on the news platform, and the reporters are penalized for publishing fake news.

%% file: conclusion.tex
\section{Conclusion and Future Work}
\label{sec:conc}

In this paper, a novel idea of the \emph{Deep Learning-based Hybrid Model to Detect Fake News using Blockchain} is presented. There are several methods experimented within the existing literature. However, the blend of deep learning on a well-established platform like blockchain can provide correctness, integrity with less human efforts, and optimal use of historical experiences. The proposed hybrid model provides various features that are not provided by deep learning and blockchain models independently as experimented within present studies. The initial implementation of the proposed model is done using the Ethereum blockchain. The future work will focus on two different implementations of the DeHiDe model using public and private blockchains. Further testing on various fake news benchmarks and identifying more practical values for various constraints, weights, and thresholds left as future work. The proposed DeHiDe model is expected to outperform state-of-art models.

%% file: main-DeHiDe.bbl
\begin{thebibliography}{10}
\providecommand{\url}[1]{\texttt{#1}}
\providecommand{\urlprefix}{URL }
\providecommand{\doi}[1]{https://doi.org/#1}

\bibitem{agarwal2020fake}
Agarwal, A., Mittal, M., Pathak, A., Goyal, L.M.: Fake news detection using a
  blend of neural networks: An application of deep learning. SN Computer
  Science  \textbf{1}, ~1--9 (2020)

\bibitem{Agarwal+Poster:ICDCN:2021}
Agrawal, P., Anjana, P.S., Peri, S.: Dehide: Deep learning based hybrid model
  to detect fake news using blockchain. In: 22nd International Conference on
  Distributed Computing and Networking 2021 (ICDCN’21). ACM (January 2021).
  \doi{10.1145/3427796.3430003}

\bibitem{anjana:ObjSC:Netys:2020}
Anjana, P.S., Attiya, H., Kumari, S., Peri, S., Somani, A.: Efficient
  concurrent execution of smart contracts in blockchains using object-based
  transactional memory. In: 8th International Conference on Networked Systems.
  NETYS'20, Springer (2020)

\bibitem{Anjana+:CESC:PDP:2019}
Anjana, P.S., Kumari, S., Peri, S., Rathor, S., Somani, A.: An efficient
  framework for optimistic concurrent execution of smart contracts. In: 2019
  27th Euromicro International Conference on Parallel, Distributed and
  Network-Based Processing (PDP'19). pp. 83--92. IEEE (Feb 2019).
  \doi{10.1109/EMPDP.2019.8671637}

\bibitem{arquam2018blockchain}
Arquam, M., Singh, A., Sharma, R.: A blockchain based secure and trusted
  framework for information propagation on online social networks. arXiv
  preprint arXiv:1812.10508  (2018)

\bibitem{Bajaj2017TP}
Bajaj, S.: The pope has a new baby! fake news detection using deep learning.
  \url{web.stanford.edu/class/archive/cs/cs224n/cs224n.1174/reports/2710385.pdf}
  (2017)

\bibitem{balouchestani2019sanub}
Balouchestani, A., Mahdavi, M., Hallaj, Y., Javdani, D.: Sanub: A new method
  for sharing and analyzing news using blockchain. In: 2019 16th International
  ISC (Iranian Society of Cryptology) Conference on Information Security and
  Cryptology (ISCISC). pp. 139--143. IEEE (2019)

\bibitem{chen2018towards}
Chen, Y., Li, Q., Wang, H.: Towards trusted social networks with blockchain
  technology. arXiv preprint arXiv:1801.02796  (2018)

\bibitem{FragaDigitalDeception2020}
{Fraga-Lamas}, P., {Fernández-Caramés}, T.M.: Fake news, disinformation, and
  deepfakes: Leveraging distributed ledger technologies and blockchain to
  combat digital deception and counterfeit reality. IT Professional
  \textbf{22}(2),  53--59 (March 2020). \doi{10.1109/MITP.2020.2977589}

\bibitem{islam2020newstradcoin}
Islam, A., Kader, M.F., Islam, M.M., Shin, S.Y.: Newstradcoin: A blockchain
  based privacy preserving secure news trading network. In: IC-BCT 2019, pp.
  21--32. Springer (2020)

\bibitem{paul2019fake}
Paul, S., Joy, J.I., Sarker, S., Ahmed, S., Das, A.K., et~al.: Fake news
  detection in social media using blockchain. In: 2019 7th International
  Conference on Smart Computing \& Communications (ICSCC). pp.~1--5. IEEE
  (2019)

\bibitem{qayyum2019using}
Qayyum, A., Qadir, J., Janjua, M.U., Sher, F.: Using blockchain to rein in the
  new post-truth world and check the spread of fake news. IT Professional
  \textbf{21}(4),  16--24 (2019)

\bibitem{shang2018tracing}
Shang, W., Liu, M., Lin, W., Jia, M.: Tracing the source of news based on
  blockchain. In: 2018 IEEE/ACIS 17th International Conference on Computer and
  Information Science (ICIS). pp. 377--381. IEEE (2018)

\bibitem{ZhangFakeDetector2020}
{Zhang}, J., {Dong}, B., {Yu}, P.S.: Fakedetector: Effective fake news
  detection with deep diffusive neural network. In: 36th International
  Conference on Data Engineering (ICDE'20). pp. 1826--1829 (2020)

\end{thebibliography}
